\documentclass{article}

\usepackage[dblblindworkshop, final]{neurips_2026}
\workshoptitle{Foundation Models for Temporal Systems: From Forecasting to World Modeling}
\setcitestyle{numbers,square,sort&compress}

\usepackage{fontspec}
\usepackage{hyperref}       % hyperlinks
\hypersetup{colorlinks = true, urlcolor = ciTeal, linkcolor = ciTeal, citecolor=ciTeal}

\usepackage{url}            % simple URL typesetting
\usepackage{booktabs}       % professional-quality tables
\usepackage{amsfonts}       % blackboard math symbols
\usepackage{nicefrac}       % compact symbols for 1/2, etc.
\usepackage{microtype}      % microtypography
\usepackage{xcolor}         % colors

\usepackage{comment}

\usepackage{graphicx}
\usepackage{amsmath, amssymb, amsthm}

\usepackage{fontspec}

\newcommand{\aptos}{\rmfamily}

\title{%Pareto-efficient parallel-in-time time series foundation models

Time Flies When You Parallelise: Pareto-efficient Time Series Foundation Models with Linear Recurrences}

\author{%
  Torben~Berndt\textsuperscript{1,2,*} \quad
  Lingyi~Yang\textsuperscript{1,3,*} \quad
  Federico~Nardi\textsuperscript{1,*} \quad
  Yi~Heng~Lim\textsuperscript{1,*} \quad
  AJ~Shears\textsuperscript{1} \quad
  Benjamin~Walker\textsuperscript{1} \\
  \textsuperscript{1}Continuous Intelligence Ltd \quad
  \textsuperscript{2}Heidelberg Institute for Theoretical Studies \\
  \textsuperscript{3}School of Mathematical Sciences, University of Nottingham \\
}

\usepackage{graphicx}
\usepackage{tcolorbox}
\usepackage{fontawesome5}
\tcbuselibrary{skins}
\definecolor{ciNavy}{HTML}{173446}     % --background
\definecolor{ciText}{HTML}{FAFBFB}     % --text
\definecolor{ciMuted}{HTML}{96A3A9}    % --muted
\definecolor{ciTeal}{HTML}{65BBAA}     % --accent (official teal)
\colorlet{ciLine}{ciText!12!ciNavy}    % --line (#fafbfb1f) as it appears on the navy background
\definecolor{ciHighlight}{HTML}{E34F5A}% secondary highlight colour

\AtBeginDocument{%
  \newgeometry{textwidth=6.5in, textheight=9in, left=1in, top=1in,
    headheight=12pt, headsep=16pt, footskip=30pt}%
  \setlength{\headwidth}{\textwidth}%
}
\usepackage{fancyhdr}
\newcommand{\ciSans}{\sffamily}

\newtcbox{\ciPill}{
  on line,
  enhanced,
  frame hidden,
  interior hidden,
  borderline={0.5pt}{0pt}{ciMuted},
  arc=1.6mm,
  left=2.2mm,
  right=2.6mm,
  top=1.1mm,
  bottom=1.1mm,
  boxsep=0pt
}
\newcommand{\ciLinkButton}[4]{\href{#1}{\ciPill{\ciSans\scriptsize
  \raisebox{-0.2em}{\makebox[0pt][l]{\rule[-0.1em]{0pt}{1.3em}}#2}\hspace{1.5mm}{\bfseries\color{ciText}#3}%
  \if\relax\detokenize{#4}\relax\else\hspace{1.4mm}{\color{ciMuted}#4}\fi}}}

\usepackage{hoverconfetti}

\begin{document}

% ---------- Continuous Intelligence technical-report title block ----------
\thispagestyle{empty}
\vspace*{-8mm}%
% \begin{tcolorbox}[enhanced, colback=ciNavy, boxrule=0pt, frame hidden, arc=2.5mm,
\begin{tcolorbox}[enhanced, colback=ciNavy, opacityback=0.90, boxrule=0pt, frame hidden, arc=2.5mm,
  left=7mm, right=7mm, top=5mm, bottom=5.5mm]
  \ciSans\raggedright\color{ciText}
  % masthead
  \raisebox{-0.5\height}{\includegraphics[height=11mm]{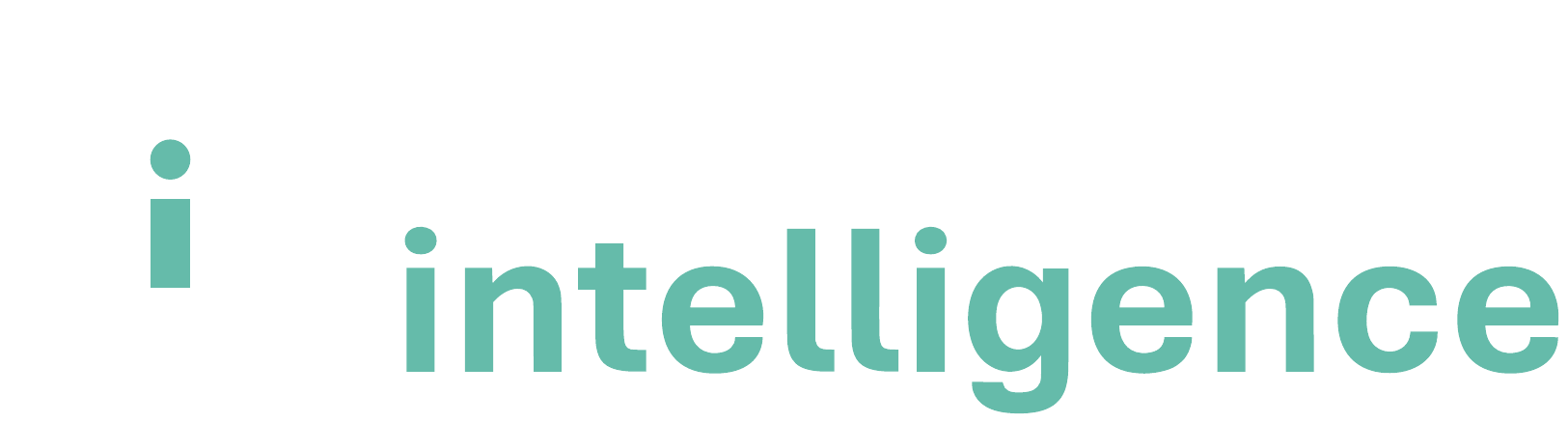}}\hfill
  {\scriptsize\color{ciHighlight}\textls[120]{TECHNICAL REPORT}%
    \hspace{2.2mm}{\color{ciHighlight}\rule[-0.15em]{0.4pt}{0.9em}}\hspace{2.2mm}%
    \textls[120]{12 NOVEMBER 2026}}\par
  \vspace{3mm}%
  {\color{ciLine}\rule{\linewidth}{0.4pt}}\par
  \vspace{4.5mm}%
  % venue
  {\footnotesize\color{ciMuted}NeurIPS 2026 Workshop on \textit{Foundation Models for Temporal Systems: From Forecasting to World Modeling}\par}
  \vspace{3mm}%
  % title
  {\fontsize{15.5}{19.5}\selectfont\bfseries
    Time Flies When You Parallelise:\\ Pareto-efficient Time Series Foundation Models with Linear Recurrences\par}
  \vspace{3mm}%
  {\color{ciTeal}\rule{14mm}{1.2pt}}\par
  \vspace{3mm}%
  % authors
  {\normalsize
    Torben~Berndt\textsuperscript{\color{ciTeal}1,2,*}\hspace{2mm}%
    Lingyi~Yang\textsuperscript{\color{ciTeal}1,3,*}\hspace{2mm}%
    Federico Nardi\textsuperscript{\color{ciTeal}1}\hspace{2mm}%
    Yi Heng Lim\textsuperscript{\color{ciTeal}1}\hspace{2mm}%
    AJ~Shears\textsuperscript{\color{ciTeal}1}\hspace{2mm}%
    Benjamin~Walker\textsuperscript{\color{ciTeal}1}\par}
  \vspace{1.5mm}%
  {\footnotesize\linespread{1.1}\selectfont\color{ciMuted}
    \mbox{\textsuperscript{1}Continuous Intelligence Limited}\hspace{3.5mm}%
    \mbox{\textsuperscript{2}Heidelberg Institute for Theoretical Studies}\hspace{3.5mm}%
    \mbox{\textsuperscript{3}University of Nottingham}\par
    \vspace{0.5mm}\textsuperscript{*}Equal contribution\par}
  \vspace{4.5mm}%
  % abstract
  {\small\bfseries\color{ciTeal}\textls[120]{ABSTRACT}\par}
  \vspace{1.5mm}%
  {\aptos\small
     \rightskip=0pt\relax
     \parfillskip=0pt plus 1fil\relax
     \linespread{1.04}\selectfont
     \color{ciText}
    %Time series foundation models (TSFMs) for multivariate zero-shot forecasting have recently gained significant traction. Pre-trained on large and diverse time series corpora, these models aim to generalise across forecasting tasks without task-specific training. Existing approaches process the temporal dimension using either Transformers, which incur quadratic complexity in sequence length, or nonlinear recurrent neural networks (RNNs), which cannot be parallelised over time. In this work, we introduce the first frontier time series foundation model based on a linear RNN. Specifically, we present ci-proto, a family of Mamba-3-based TSFMs that lie on the Pareto frontier of fev-bench, demonstrating that linear RNNs can achieve Pareto state-of-the-art performance in zero-shot forecasting.

    Time series foundation models have demonstrated strong zero-shot forecasting performance, but computational efficiency remains a central architectural challenge. Temporal self-attention incurs quadratic cost in the number of temporal tokens, while nonlinear recurrent state transitions restrict parallel computation along the temporal dimension. We introduce 
    % \confetti{\textbf{\textcolor{ciTeal}{ci-proto}}}
    {\textbf{\textcolor{ciTeal}{ci-proto}}}
    , a family of time series foundation models based on Mamba-3, whose linear state recurrences support chunk-wise parallel-in-time computation during training, and inference costs that scale linearly with sequence length. The architecture alternates recurrent temporal processing with cross-variate attention, supporting multivariate probabilistic forecasting with past-only and future-known covariates. 
    Across fev-bench's 100 forecasting tasks, ci-proto XL sets a new state of the art among time series foundation models. The five ci-proto variants also make up the entire Pareto frontier for mean scaled quantile loss and mean absolute scaled error against both parameter count and median runtime.
  \par}
  \vspace{4mm}%
  \ciLinkButton{https://github.com/continuousintelligence/ci}%
    {{\color{ciText}\normalsize\faGithub}}{Code}{GitHub}%
  \hspace{2.5mm}%
  \ciLinkButton{https://huggingface.co/continuousintelligence/ci-proto}%
    {\includegraphics[height=1.15em]{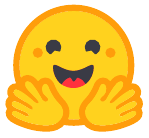}}{Models}{Hugging Face}%
  \hspace{2.5mm}%
  \ciLinkButton{https://www.continuousintelligence.co.uk/}%
    {{\color{ciTeal}\normalsize\faGlobe}}{Website}{continuousintelligence.co.uk}%
  \par
\end{tcolorbox}
\vspace{3mm}

\begin{figure}[h]
    \centering
    \includegraphics[width=0.95\linewidth]{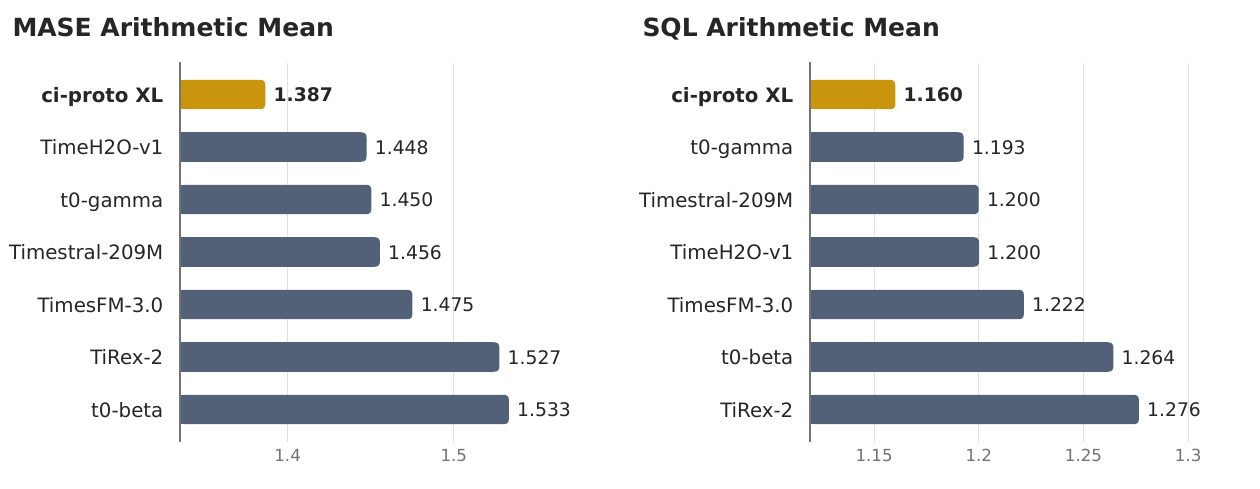}
    \caption{
        \textbf{Arithmetic mean errors of ci-proto on fev-bench.} Top seven models on the official \texttt{fev-bench} leaderboard~\cite{shchur2026fevbench}. The left panel shows the mean absolute scaled error (MASE) for point forecasting, and the right panel shows the scaled quantile loss (SQL) for probabilistic forecasting. Lower values indicate better performance in both panels.
    }
    \label{fig:fev_leaderboard_mean}
\end{figure}

\section{Introduction}

Large-scale pre-training has shifted machine learning from bespoke,
task-specific models to foundation models that transfer across tasks
and datasets, driving advances in
language~\citep{devlin2019bert,brown2020gpt3},
vision~\citep{badford2021clip,caron2021dino,oquab2024dinov2},
and speech~\citep{radford2023whisper}.
A similar shift is underway in time series forecasting.
Time series foundation models (TSFMs) learn from diverse corpora
to forecast unseen datasets without task-specific training.
Early models such as TimeGPT~\citep{garza2023timegpt},
TimesFM~\citep{das2024timesfm}, and
Chronos~\citep{ansari2024chronos} established this paradigm,
with subsequent work extending it to multivariate forecasting
and the use of covariates~\citep{woo2024morai1,ansari2025chronos,podest2026tirex2}.

Recent TSFMs have reached the billion-parameter scale, making computational efficiency a key consideration \citep{shi2025timemoe,khwaja2026toto2}.
A central architectural choice is how to model temporal dependencies.
Most TSFMs use the Transformer architecture, whose self-attention permits parallel
computation across temporal positions but incurs quadratic cost in
the number of tokens.
Recurrent architectures avoid this quadratic temporal cost, but
not all recurrences can be efficiently parallelised.
TiRex~\citep{auer2025tirex}, for example, uses nonlinear sLSTM
recurrences, while TiRex-2~\citep{podest2026tirex2} combines nonlinear
sLSTM and linear mLSTM recurrences.
The sLSTM layers require each recurrent state to be computed before
the next can be evaluated, introducing a sequential bottleneck during
training and context processing~\citep{beck2024xlstm}.

In contrast, linear recurrences have state updates that  
are linear in the previous hidden state, which
allows successive updates to be composed
without explicitly evaluating the intermediate hidden states \citep{gu2024mamba}.
This enables parallel scan algorithms~\citep{martin2018parallelizing,gu2024mamba}
and hardware-efficient chunk-wise algorithms that combine parallel
matrix multiplications within chunks with recurrent propagation
between them~\citep{yang2024parallelizing,dao2024transformers,lahoti2026mamba}.
Linear recurrent models can therefore support parallel-in-time
training and context processing while retaining recurrent inference
whose cost scales linearly with sequence length.

However, existing TSFMs with entirely linear recurrent temporal backbones,
including Mamba4Cast~\citep{bhethanabhotla2024mamba4cast},
FlowState~\citep{graf2026flowstate}, and TempoPFN~\citep{moroshan2026tempopfn}, are 
limited to processing individual series without cross-variate interaction.
We address this gap with {ci-proto}, a family of TSFMs
that alternate Mamba-3 temporal processing~\citep{lahoti2026mamba}
with cross-variate attention.
Mamba-3 independently processes the temporal sequence of each
variate, while attention exchanges information between variates at
each temporal position.
To the best of our knowledge, ci-proto is the first multivariate
TSFM with an entirely linear recurrent temporal backbone.
The architecture supports probabilistic forecasting with past-only
and future-known covariates, chunk-wise parallel-in-time training,
and inference costs that scale linearly with sequence length.

Our choice of Mamba-3 is motivated by both computational efficiency
and the expressivity of its recurrent dynamics.
Restrictions on state-transition matrices, including diagonal
structure and constraints on their eigenvalues, can limit the
expressivity and state-tracking capabilities of linear recurrent
models~\citep{merrill2024illusion,walker2025structured,grazzi2025unlocking}.
Mamba-3 enriches these dynamics through input-dependent
$2 \times 2$ rotation matrices, improving state tracking relative
to Mamba-2 while retaining hardware-efficient chunk-wise
computation~\citep{lahoti2026mamba}.

Across the $100$ forecasting tasks in \texttt{fev-bench}, ci-proto XL sets a new 
state of the art among time series foundation models. The five ci-proto variants
also make up the entire Pareto frontier for mean scaled quantile loss and mean 
absolute scaled error against both parameter count and median runtime. These 
results demonstrate that state-of-the-art multivariate zero-shot forecasting can 
be achieved without temporal self-attention or nonlinear state recurrences.

\section{ci-proto}

\subsection{Problem setting}
We consider $d$ partially observed variables over a continuous time interval $[0,T]$. 
For each variable $i \in \{1,\ldots,d\}$, we observe a path
$X^i : J_i \to \mathbb{R}$, where $J_i \subseteq [0,T]$ is a union of
observation intervals. 
The observation sets $J_i$ need not be connected and may differ across variables. 
Writing $X=(X^1,\ldots,X^d)$, we seek to model the
conditional distribution $\mathbb{P}(Y \vert X)$ of a complete path
$Y:[0,T]\to\mathbb{R}^d$ consistent with the observations, that is,
$Y^i|_{J_i}=X^i$ for every $i$. 

We focus on forecasting as a special case of this formulation.
Let $0 \leq T' < T$ denote the forecast origin and $[T',T]$ the
forecast interval. Variables are assigned one of three roles.
\emph{Targets} are variables to be forecast, with observations
restricted to the period before the forecast origin, so that
$J_i \subseteq [0,T')$. \emph{Past-only covariates} provide
additional information observed before the forecast origin,
again with $J_i \subseteq [0,T')$. \emph{Future-known covariates}
provide information within the forecast interval, with
$J_i \cap [T',T] \neq \emptyset$; examples include scheduled
promotions and calendar features such as holidays.
The forecasting objective is to model the conditional distribution
of the targets over $[T',T]$.
% given all available target and covariate observations.

\subsection{Architecture}

For the architecture considered here, we work with a finite number
of regularly spaced samples. We group these samples into $L$
temporal patches and represent the patched inputs as a sequence
of hidden representations
\(
    H^{(0)} : \{1,\ldots,L\} \to \mathbb{R}^{d \times D},
\)
where $d$ is the number of variates, $D$ is the hidden dimension
and $H_i^{(0)}(k)$ represents patch $k$ of variate $i$. For more details on the patching, see Appendix~\ref{app:patching}. 
To capture both temporal dependencies and cross-variate
relationships, we alternate temporal processing with
cross-variate mixing, following
\citep{ansari2025chronos,podest2026tirex2}.
Each block applies a time mixer $\mathcal{T}_n$, which processes
each variate independently along the temporal axis, followed by
a variate mixer $\mathcal{V}_n$, which exchanges information
between variates at each temporal position:
\[
    H^{(2n+1)} = \mathcal{T}_n\bigl(H^{(2n)}\bigr),
    \qquad
    H^{(2n+2)} = \mathcal{V}_n\bigl(H^{(2n+1)}\bigr).
\]

\paragraph{Variate mixer.}
The variate mixer applies attention across variates at each patch position:
\[
    \bigl[\mathcal{V}_n(H)\bigr](k)
    = \operatorname{GroupAttn}_n\bigl(H(k)\bigr),
    \qquad k=1,\ldots,L.
\]
We use TiRex-2's asymmetric group attention
\citep{podest2026tirex2}, building on
\citep{ansari2025chronos,cohen2026toto1,feng2024general}:
past-only and future-known covariates attend to one another,
while targets attend to both types of covariates, but not vice versa.

\begin{figure}
    \centering
    \includegraphics[width=0.9\linewidth]{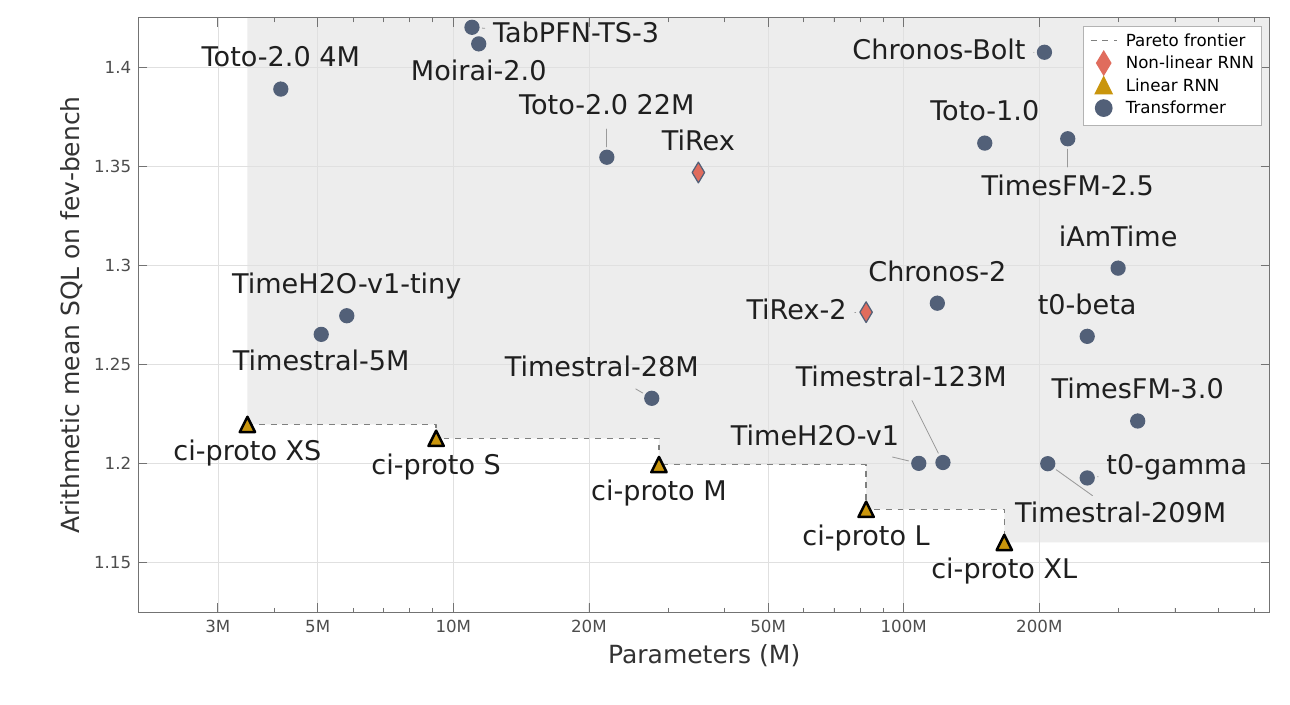}
    \caption{
        \textbf{Mean scaled quantile loss (SQL) on fev-bench versus parameter count}.
        Lower SQL is better. The horizontal axis is in logarithmic scale. The dashed line marks the Pareto frontier formed by the five ci-proto variants.
    }
    \label{fig:pareto_frontier_params}
\end{figure}

\begin{figure}
    \centering
    \includegraphics[width=0.9\linewidth]{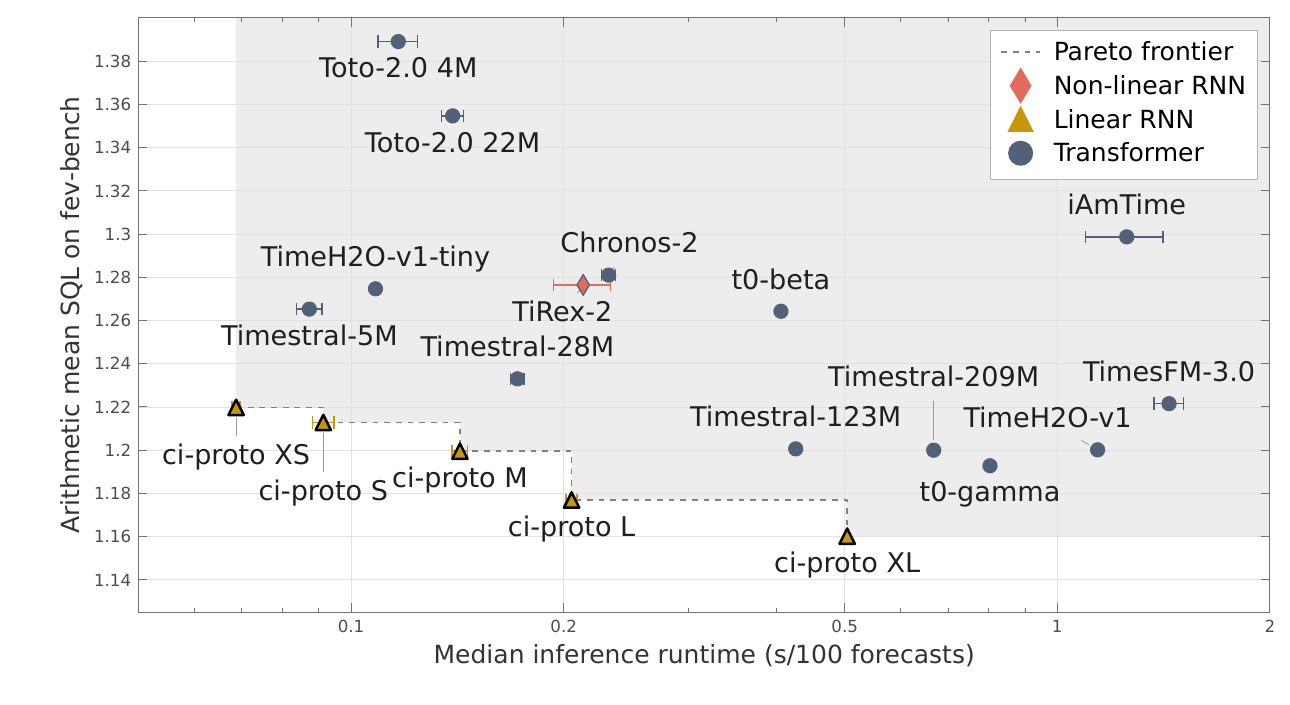}
    \caption{
        \textbf{Mean scaled quantile loss (SQL) on fev-bench versus median runtime per 100 forecasts for selected leaderboard models.} We benchmark all models on identical
        hardware and under the same execution settings, since self-reported
        leaderboard timings are not directly comparable. We report the
        mean and standard deviation across five runs. Lower SQL is better. The horizontal axis is in logarithmic scale. The dashed line marks the Pareto frontier formed by the five ci-proto
        variants.
    }
    \label{fig:pareto_frontier_runtime}
\end{figure}

\paragraph{Time mixer.}
The time mixer applies Mamba-3~\citep{lahoti2026mamba}
independently to each variate's patch sequence, with parameters
shared across variates:
\[
    \bigl[\mathcal{T}_n(H)\bigr]_i
    = \operatorname{Mamba3}_n
      \bigl(H_i(1),\ldots,H_i(L)\bigr),
    \qquad i=1,\ldots,d.
\]
The time mixer processes all three variate types causally, with its state evolving along the discrete patch index $k$. Future-known covariates are additionally processed backwards in time using the same parameters. For information on the chosen hyperparameters, see Appendix \ref{app:model_config}.

\paragraph{Forecasting head and loss.}
For each target variate $i$ and time step in patch $k+1$,
$9$ linear heads predict equally-spaced quantiles $0.1, \dots, 0.9$ from the final-layer
representation $H_i(k)$.
We minimise the pinball loss
\citep{koenker1978pinball}, averaged over quantile levels
and observed target values. For more information, see Appendix \ref{app:quantile_and_objective}.

\paragraph{Training dataset.}
Following TiRex-2 \citep{podest2026tirex2}, we construct a pre-training
corpus from subsets of the Chronos training data \citep{ansari2024chronos}
and the GIFT-Eval pre-training dataset
\citep{aksu2024giftevalbenchmarkgeneraltime} (filtered for leakage in \texttt{fev-bench}).
We add synthetic sequences generated by Gaussian processes
\citep{ansari2024chronos,auer2025tirex} and by generators of multivariate
processes with covariate effects.
During training, we randomly sample univariate sequences from this corpus and combine them into correlated multivariate sequences using either identity mappings, functional coupling, linear mixing, co-integration, or linear and non-linear structural causal models.
Finally, we enrich these examples with binary covariates or
calendar features and assign each variate a role as a forecasting
target, past-only covariate, or future-known covariate.
After pre-training, we post-train on additional publicly available
real-world datasets and synthetic data, also filtered for leakage against
the evaluation benchmarks, mixed with subsets of the earlier training data.
For more detail on the dataset composition, see Appendix \ref{app:training}.

\section{Experiments}
\label{sec:experiments}

\subsection{Standard forecasting benchmarks}
\label{sec:standard_forecasting}

% In this section, we evaluate whether linear recurrent temporal processing provides a competitive accuracy-efficiency trade-off for multivariate zero-shot forecasting.

\paragraph{Benchmark and models.}
We evaluate five ci-proto variants (XS, S, M, L, and XL) on the
$100$ forecasting tasks in \texttt{fev-bench}~\citep{shchur2026fevbench},
covering diverse domains and forecasting settings, including
tasks with past-only and future-known covariates.
We compare against TabPFN-TS-3~\citep{hoo2026tables, grinsztajn2026tabpfn3},
Moirai-2.0~\citep{liu2025moirai2},
TiRex~\citep{auer2025tirex} and 
TiRex-2~\citep{podest2026tirex2},
Chronos-2~\citep{ansari2025chronos} and Chronos-Bolt~\citep{ansari2024chronosbolt},
TimesFM-2.5 and TimesFM-3~\citep{das2024timesfm, jain2026timesfm3}, Toto-1.0~\citep{cohen2026toto1} and Toto-2.0~\citep{khwaja2026toto2}, the Timestral family (5M, 28M, 123M, and 209M)~\citep{mistral2026timestral}, TimeH2O-v1-tiny and TimeH2O-v1~\citep{h2o2026timeh2o}, the $t_0$ family (beta and gamma)~\citep{meyer2026t0, tfc2026t0gamma}, and iAmTime~\citep{saha2026iamtime}.
Baseline forecasting scores are taken from the official leaderboard,
while runtimes are measured by us on identical hardware and under
the same execution settings.

\paragraph{Test-time adaptation.}
Inspired by TiRex-2~\citep{podest2026tirex2}, TimesFM-3~\citep{jain2026timesfm3}, the $t_0$
family~\citep{meyer2026t0}, Timestral~\citep{mistral2026timestral},
and TimeH2O~\citep{h2o2026timeh2o}, we apply a
model-free test-time adaptation around the forward pass. Specifically,
strongly trending targets are differenced before forecasting and
integrated back afterwards, crossing quantiles are sorted, and each
forecast is averaged with the mapped-back forecast of the negated
series. Two constants are fitted on the
\texttt{fev-bench} validation split and then frozen for the test
evaluation (Appendix~\ref{app:tta}).

\paragraph{Metrics and efficiency.}
We evaluate point and probabilistic forecasting using mean absolute scaled error (MASE) and scaled quantile loss (SQL), respectively~\citep{shchur2026fevbench}. For each metric, we report arithmetic means across series within each task and across tasks, together with a skill score relative to the seasonal naive baseline:
\[
    \operatorname{SkillScore}_{m}
    = 1 - \left(
        \prod_{t=1}^{T}
        \frac{m_t^{\mathrm{model}}}{m_t^{\mathrm{seasonal\ naive}}}
    \right)^{1/T},
    \qquad m \in \{\mathrm{MASE}, \mathrm{SQL}\},
\]
where \(T\) is the number of tasks and \(m_t^{\mathrm{model}}\) and \(m_t^{\mathrm{seasonal\ naive}}\) denote the corresponding errors on task \(t\). We consider these metrics both individually and in relation to parameter count and runtime per $100$ forecasts.

\paragraph{Forecasting accuracy and efficiency.}

\begin{figure}
    \centering
    \includegraphics[width=0.9\linewidth]{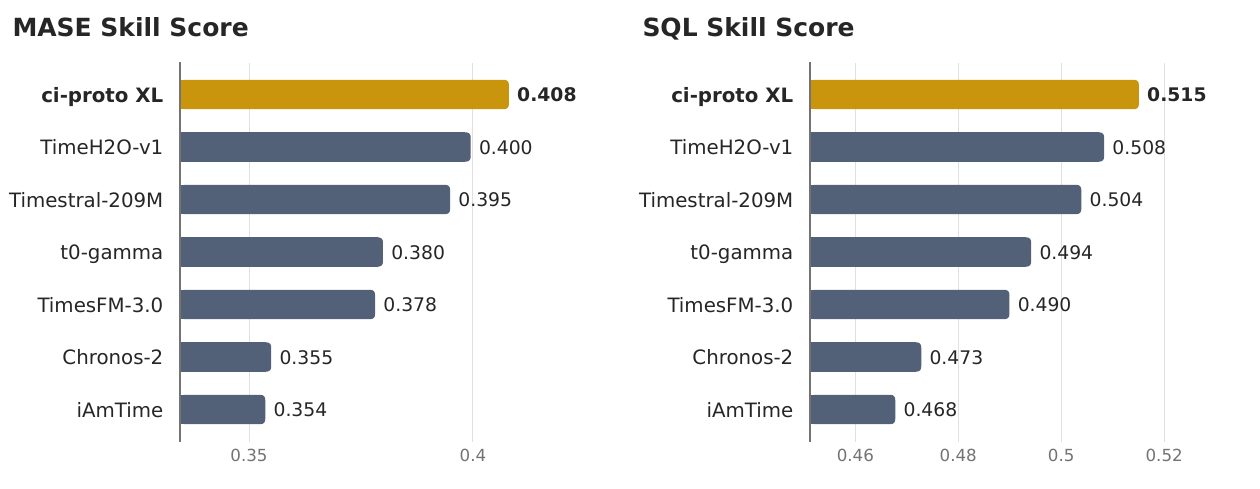}
    \caption{
       \textbf{Skill scores of ci-proto on fev-bench.} Skill score relative to the seasonal naive baseline for the top seven models on the official \texttt{fev-bench} leaderboard~\cite{shchur2026fevbench}. The left panel shows skill score based on the mean absolute scaled error (MASE) for point forecasting, and the right panel shows it based on the scaled quantile loss (SQL) for probabilistic forecasting. Higher values indicate better performance in both panels.
    }
    \label{fig:fev_leaderboard_skill}
\end{figure}

Figures~\ref{fig:fev_leaderboard_mean} and~\ref{fig:fev_leaderboard_skill} show that our largest model, ci-proto~XL, achieves the lowest arithmetic mean errors and the highest skill scores for both MASE and SQL on \texttt{fev-bench}. Figures~\ref{fig:pareto_frontier_params} and~\ref{fig:pareto_frontier_runtime} further show that the five ci-proto variants form the Pareto frontiers of mean SQL versus parameter count and measured runtime per $100$ forecasts. Furthermore, our smallest model, ci-proto~XS, achieves lower mean SQL than Timestral-28M and TimesFM-3, while using substantially fewer parameters at a faster run time. 

These results extend to point forecasting. Figures~\ref{fig:pareto-mase-params} and~\ref{fig:pareto-mase-runtime} (Appendix ~\ref{app:mase_skill}) show that the five ci-proto variants form the Pareto frontiers of mean MASE versus parameter count and median runtime. When performance is measured using MASE and SQL skill scores, all five variants remain on the corresponding Pareto frontiers (Appendix~\ref{app:mase_skill}). See Appendix~\ref{app:evaluation} for more detail on how to interpret Pareto frontiers and our hardware and runtime specifications.

\subsection{Long-context forecasting}
\label{sec:long_context}

Linear recurrences let ci-proto read histories far longer than the
$16{,}384$-step limit used on \texttt{fev-bench}, at a cost linear in their length. 
To test how well our model is able to utilise long contexts, we fine-tune ci-proto~XL on
contexts of up to $1{,}048{,}576$ steps. The data include synthetic
series with future-known binary covariates that affect target behaviour but are activated only rarely within the context (Appendix~\ref{app:long_range}).
We then forecast one-minute household electricity
consumption~\citep{hebrail2006household} with a scheduled power outage in the
horizon. The only precedent is a single earlier outage, a given number of
steps back. 

Each model receives the shortest context that includes the earlier outage, with its length rounded up to the nearest power of two and a minimum of $2,048$ steps. The context is capped at the full available history of $2,048,000$ steps.
The baselines are TiRex-2~\citep{podest2026tirex2},
Chronos-2~\citep{ansari2025chronos},
t0-gamma~\citep{tfc2026t0gamma},
TimesFM-3~\citep{jain2026timesfm3}, 
TimeH2O-v1~\citep{h2o2026timeh2o},
and Timestral-209M~\citep{mistral2026timestral}.
Each model is released with a context cap, keeping only the most recent
$2{,}048$ (TiRex-2) to $32{,}768$ (Timestral-209M) steps of a longer input.
We remove the cap so that each model reads the whole context, we refer to these as
uncapped. 

\begin{figure}
    \centering
    \includegraphics[width=0.9\linewidth]{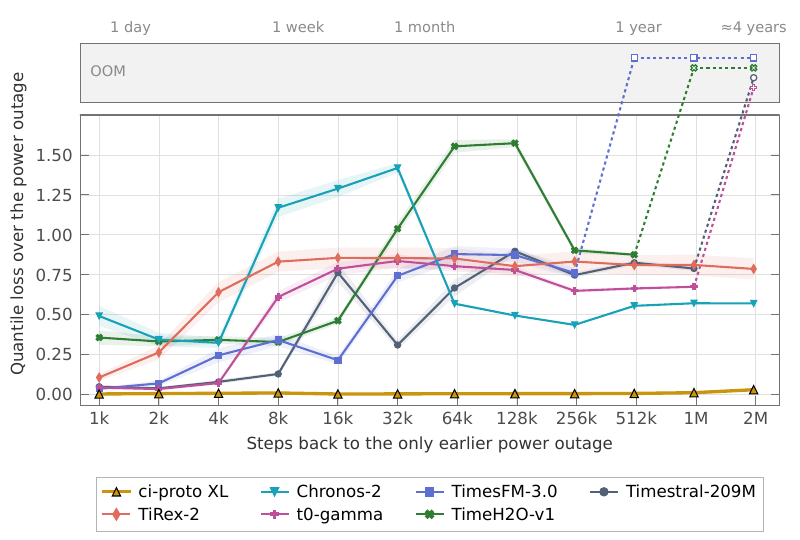}
    \caption{
        \textbf{Quantile loss over a scheduled power outage for uncapped models.} The outage has a
        single earlier occurrence at the given number of steps back as shown on the x-axis, which is logarithmic. 
        Each model receives the smallest of power-of-two context $2^k \geq 2{,}048$ steps that
        contains the earlier outage, or all $2{,}048{,}000$ steps of history if needed.
        ci-proto~XL and the uncapped baselines read the whole context, with all
        other settings matching each model's \texttt{fev-bench} submission.
        The curves show the mean and standard deviation over 40 forecast windows (lower is better). Here each step is one minute, and the top axis gives the
        steps as units of time. 
    }
    \label{fig:long_range_recall}
\end{figure}

\begin{figure}
    \centering
    \includegraphics[width=0.9\linewidth]{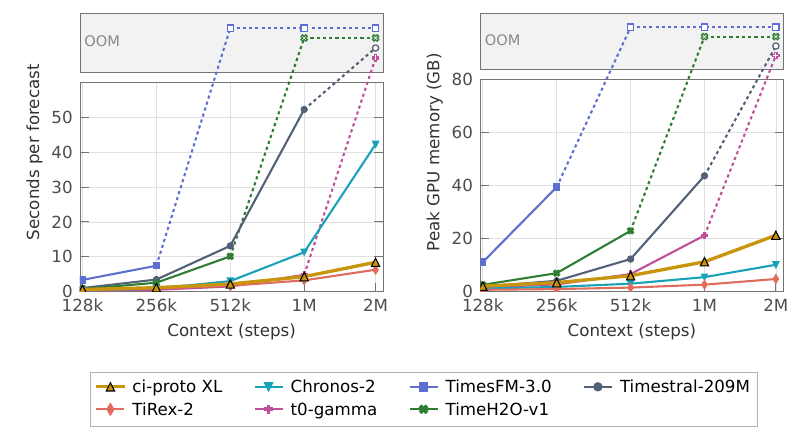}
    \caption{
        \textbf{Time and peak GPU memory per forecast for ci-proto~XL and the
        uncapped baselines.}
        The context axis is logarithmic, while the time and memory axes are
        linear and start at zero. Values are the median time and largest
        memory over 40 windows (identical to those used for Figure~\ref{fig:long_range_recall})
        on one H100 when forecasting one series at a time. All settings other than the
        removed context caps match the models' \texttt{fev-bench} submission.
    }
    \label{fig:long_range_cost}
\end{figure}

As Figure~\ref{fig:long_range_recall} shows, ci-proto~XL forecasts the outage
almost exactly from as far as $2{,}000{,}000$ steps ($3.8$ years) back.
ci-proto~XL's runtime and memory per forecast grow linearly with the
context, to $8$\,s and $21$\,GB at $2{,}048{,}000$ steps on one H100.
TiRex-2's runtime and memory grow linearly too, to $6$\,s and $5$\,GB, but it recalls the outage from
only a few thousand steps back.
Four of the five uncapped Transformers exceed the GPU memory of an
$80$\,GB H100 before two million steps, and the fifth, Chronos-2, takes $42$\,s
(Figure~\ref{fig:long_range_cost}).
Figure~\ref{fig:long_range_recall_capped} (Appendix~\ref{app:long_range}) compares ci-proto with the baselines using their
released cap. In general, we see that compared with their uncapped version, the models as released have higher quantile loss when the previous outage took place outside of their context.

\section{Conclusion}

We introduced ci-proto, a family of multivariate time series
foundation models combining Mamba-3 temporal processing
with cross-variate attention.
Its linear state recurrences enable chunk-wise parallel
computation, while supporting probabilistic forecasting
with past-only and future-known covariates.
Across fev-bench's $100$ forecasting tasks, ci-proto XL sets a new state of the art among time series foundation models. The five ci-proto variants make up the entire Pareto frontier for mean scaled quantile loss and mean absolute scaled error against both parameter count and median runtime.
Fine-tuned on long contexts, ci-proto is able to leverage information from millions of
steps back, recalling a single event from about $4$ years earlier in a
minute-resolution series, where Transformer and non-linear recurrent baselines 
cannot.
Our findings establish linear recurrent temporal backbones as a state-of-the-art architectural choice for multivariate time series foundation models, combining leading forecasting accuracy with computational efficiency.

\begin{ack}
Continuous Intelligence Limited is supported by SPRIN-D, the German Federal Agency for Breakthrough Innovation, through their Next Frontier AI Challenge.
\end{ack}

%%%%%%%%%%%%%%%%%%%%%%%%%%%%%%%%%%%%%%%%%%%%%%%%%%%%%%%%%%%%

% \appendix

% \section{Technical appendices and supplementary material}

%%%%%%%%%%%%%%%%%%%%%%%%%%%%%%%%%%%%%%%%%%%%%%%%%%%%%%%%%%%%

\newpage

\bibliographystyle{plainnat}
\bibliography{main/main}

\newpage

\appendix

\appendix

\section{Model configuration and forecasting objective}
\label{app:architecture}

\subsection{Model configurations}
\label{app:model_config}

Table~\ref{tab:model-configurations} reports the configurations
of the five ci-proto variants. We count one block as a time mixer
followed by a variate mixer, as defined in the main text.
The temporal parameters are shared across variates, so increasing
the number of input variates does not require a separate
set of temporal-model parameters for each variate.

\subsection{Normalisation}

We normalise each variate independently using its mean $\mu$ and standard deviation $\sigma$ over observed inputs:
\[
  \tilde{x}_t =
  \operatorname{arcsinh}\!\left(
      \frac{x_t-\mu}{\max(\sigma,\epsilon)}
  \right),
  \qquad \epsilon=10^{-8}.
\]
Binary variates are exempt from normalisation in order to retain their $\{0,1\}$ encoding.

\subsection{Patching}
\label{app:patching}

Context sequences are left-padded with missing values to the nearest power of two and divided into non-overlapping patches of length $P=32$.
We define an observation mask $m_t=\mathbf{1}\{x_t\text{ observed}\}$ and replace missing and padded values by zero after normalisation.
Each patch is concatenated with its mask,
\[
  \mathbf{z}_\ell =
  [\tilde{\mathbf{x}}_\ell;\mathbf{m}_\ell]
  \in\mathbb{R}^{2P},
\]
and embedded using a shared residual MLP:
\[
  \mathbf{h}_\ell =
  W_2\phi(W_1\mathbf{z}_\ell+\mathbf{b}_1)+\mathbf{b}_2
  +W_r\mathbf{z}_\ell+\mathbf{b}_r
  \in\mathbb{R}^{D}.
\]
Boolean masks encode target, past-only, and known-future covariate roles. Only known-future covariates receive bidirectional temporal processing.
Variate attention is restricted to each sample, excludes variates introduced purely for padding, and prevents covariate queries from attending to target keys.

\begin{table}[b]
    \centering
    \small
    \caption{\textbf{ci-proto model configurations.} Parameter counts are
    reported in millions.}
    \label{tab:model-configurations}
    \begin{tabular}{lrrrrr}
        \toprule
        Configuration & XS & S & M & L & XL \\
        \midrule
        Number of blocks
        & 6 & 8 & 10 & 12 & 16 \\
        Hidden dimension $D$
        & 128 & 192 & 320 & 512 & 640 \\
        Patch length $P$
        & 32 & 32 & 32 & 32 & 32 \\
        Mamba-3 state dimension
        & 128 & 128 & 128 & 128 & 128 \\
        Mamba-3 head dimension
        & 64 & 64 & 64 & 64 & 64 \\
        Variate-attention heads
        & 4 & 4 & 4 & 4 & 4 \\
        Parameters
        & 3.49 & 9.16 & 28.61 & 82.51 & 167.19 \\
        \bottomrule
    \end{tabular}
\end{table}

\subsection{Quantile predictions and training objective}
\label{app:quantile_and_objective}

Let $\mathcal{Q}\subset(0,1)$ denote the set of $9$
predicted quantile levels, and write
$\widehat{y}^{(q)}_{i,t}$ for the prediction at level $q$
for target variate $i$ and time step $t$.
For each time step in patch $k+1$, these predictions are
obtained from linear heads applied to the final-layer
representation $H_i(k)$.

For residual $u$, the pinball loss at quantile level $q$ is
\begin{equation}
    \rho_q(u)
    =
    \begin{cases}
        q u, & u \geq 0, \\
        (q-1)u, & u < 0.
    \end{cases}
\end{equation}
Writing $\mathcal{O}$ for the observed target positions
included in the loss, the objective is
\begin{equation}
    \mathcal{L}
    =
    \frac{1}{|\mathcal{O}|\,|\mathcal{Q}|}
    \sum_{(i,t)\in\mathcal{O}}
    \sum_{q\in\mathcal{Q}}
    \rho_q\!\left(
        y_{i,t}-\widehat{y}^{(q)}_{i,t}
    \right).
    \label{eq:appendix-pinball}
\end{equation}
This is the pinball objective~\citep{koenker1978pinball} described in the main
text. Unobserved target values are excluded from the loss.

The supervision is applied separately to each target value
and quantile level, while the representations used to form
these predictions incorporate information exchanged through
the variate mixers. Cross-variate conditioning therefore
enters through the forecasting representation rather than
through an additional cross-variate term in the loss.

\section{Pre-training and optimisation}
\label{app:training}

\subsection{Construction of pre-training examples}

As in TiRex-2~\citep{podest2026tirex2}, we train on multivariate
examples generated from a corpus of source series. The corpus contains
parts of the Chronos training data and the GIFT-Eval pre-training
dataset, filtered for leakage (in \texttt{fev-bench}). It also contains synthetic series from
Gaussian processes and from a generator of multivariate series with
covariates.

During training, sampled univariate sequences are combined
using identity mappings, functional coupling, linear mixing,
co-integration, or linear and nonlinear structural causal
models. The resulting examples are enriched with binary
covariates or calendar features, and each variate is assigned
a role: target, past-only covariate, or future-known covariate.

We follow the benchmark-specific training-data selections of TiRex-2~\citep{podest2026tirex2} 
and additionally check BOOM for exact duplicates of fev-bench's \texttt{boomlet} series. 
We hash each benchmark target channel at float32 precision and compare it against the BOOM corpus,
identifying overlapping source groups even when their names differ. We exclude all matching groups before training.

\subsection{Post-training pipeline}
\label{app:posttraining}

After pre-training, each ci-proto variant is post-trained on mixtures 
of real and synthetic data, with the context
extended from $2{,}048$ to $8{,}192$ time steps (up to $16{,}384$ for ci-proto~XL). 
Each post-training stage's length is a fixed
fraction of its base's pre-training steps, about a quarter of them in
total, and ci-proto XL uses the stage lengths of ci-proto~L. The real
data add publicly available datasets from domains such as energy,
transport, environmental sensing and economics. The synthetic data
comprise generated multivariate series with covariates and long
series that fill the extended context. Post-training also replays data from
earlier training.

Post-training uses the optimiser, schedule and precision of
Table~\ref{tab:training-settings}, with an effective batch size
of $64$ sampled groups ($32$ at the $16{,}384$-step context) and a
peak learning rate that decreases across post-training from
$2.4\times10^{-4}$ to $2.5\times10^{-5}$. We modify the
pinball loss in three ways. First, each group's loss is weighted
by the inverse of its detached value, floored at a quarter of
the batch median, so that easy and hard series contribute more
evenly. After an initial period of post-training, we also weight
the mean losses over the forecast window and over the context
equally, and train half of the groups on a shorter forecast
prefix, whose length is drawn log-uniformly with a minimum of
four steps.
  
\subsection{Test-time adaptation}
\label{app:tta}

Our test-time adaptation builds around the trend-differencing and
quantile-band postprocessor of TiRex-2~\citep{podest2026tirex2}
and combines it with categorical target encoding, calendar
covariates, related-series context, quantile sorting, sign-flip
averaging, nonnegative clipping, and integer rounding.
These operations do not update the model weights.
For each of the components, we reference corresponding operations
in the public implementations and benchmark adapters of
TiRex-2~\citep{podest2026tirex2}, TimesFM-3~\citep{jain2026timesfm3}, the $t_0$
family~\citep{meyer2026t0}, Timestral~\citep{mistral2026timestral},
and TimeH2O~\citep{h2o2026timeh2o}, where applicable.

\paragraph{Categorical target encoding.}
\emph{Also used by: Timestral and TimeH2O.}

We encode categorical covariates using smoothed target means
computed over the context steps at which each category is active.
Let \(z_t\) denote the target value at context step \(t\), averaged
across observed target variables in the multivariate case. We
denote by \(\mu\) the mean of \(z_t\) over the context region and
by \(\mathcal{I}_c\) the set of context steps at which category
\(c\) is active. The encoding of category \(c\) is then given by
\[
  e(c) = \frac{\lambda \mu + \sum_{t \in \mathcal{I}_c} z_t}
              {\lambda + |\mathcal{I}_c|},
  \qquad \text{we set }\lambda = 1.
\]

\paragraph{Calendar covariates.}
\emph{Also using derived covariates:
Chronos-2, TiRex-2, TimesFM-3, and TabPFN-TS-3.}

When timestamps and sampling frequency are available, we
generate calendar features and append them as known-future
covariates.

\paragraph{Joint forecasting in chunks.}
\emph{Also used by: $t_0$-gamma and Chronos-2.}

For tasks with independent univariate series, the enabled
chunking policy constructs joint forecasting groups.
We identify all series that share the same forecast origin,
and partition them into chunks of a configured maximum size.
All series within each chunk are treated as targets and
forecast jointly, allowing their representations to
exchange information through the variate mixers.

\paragraph{Static-group context.}
\emph{Related metadata-based grouping in Chronos-2.}

When static metadata are supplied and static grouping is
enabled, we construct additional
past-only covariates from series sharing selected metadata
values. For each selected grouping column, a target receives the
mean of the other group members' standardised histories,
excluding its own contribution. Pooling uses only observed historical values from series with the same forecast origin.

\paragraph{Correlated-neighbour context.}
\emph{ci-proto extension.}

For tasks with covariates but no static metadata, each target is
forecast jointly with seven series that share the same forecast origin and whose 
histories best correlate with the target.
These other series must have
at least $32$ overlapping observations on the latest $2{,}048$ distinct historical
timestamps.
The selected neighbours are aligned to the original target series'
timestamps and added also as target rows for a joint forward pass.
Only the original target series' forecast is retained.

\paragraph{Trend detection.}
\emph{Also used by: TiRex-2; related linear detrending in TimesFM-3.}

For each target row, we take the trailing fraction $w$ of its
available context and z-score normalise the values in that window.
We fit an ordinary-least-squares line to these standardised
values, yielding a slope $\beta_z$ and coefficient of
determination $R^2$.
For a context of length $L$, let
$n_w=\max(1,\operatorname{round}(wL))$ denote the number of
time steps in the trailing window.
A row is treated as trending if
\[
  |\beta_z|\,n_w \geq \tau
  \quad\text{and}\quad
  R^2 \geq \rho.
\]
The first condition measures the magnitude of the fitted drift
in standard-deviation units, while the second requires the
linear fit to explain a sufficient fraction of the variation.

\paragraph{Running-total detection.}
\emph{ci-proto extension.}

We extend the differencing criterion to capture cumulative series
whose changing rates can cause them to fail the linear-trend test.
For each target row, let $x_1,\ldots,x_n$ denote its observations,
and define $\Delta_j=x_{j+1}-x_j$.
A row is flagged as a running total if its observed
values are monotonic in either direction and at least half
its observed transitions are nonzero:
\[
  \left[
    \left(\forall j:\Delta_j\geq0\right)
    \;\lor\;
    \left(\forall j:\Delta_j\leq0\right)
  \right]
  \quad\text{and}\quad
  \frac{1}{n-1}\sum_{j=1}^{n-1}
    \mathbf{1}\{\Delta_j\neq0\}
  \geq \frac12.
\]
A target satisfying either this criterion or the linear-trend
criterion follows the differencing and forecast integration
procedure below.

\paragraph{Differenced targets.}
\emph{Also used by: TiRex-2.}

For prediction horizons $H\leq48$, targets selected by
either detection criterion above are replaced by their first
differences, with a zero leading step at the start of
each series, before the forward pass. Covariates are not
differenced.
The model forecasts increments for the selected targets, which
are mapped back to values as follows.
Let $H$ be the prediction horizon, $m_t$ the predicted median
increment at horizon step $t$, $o_{q,t}$ the offset of quantile
$q$ from that median increment, and $y_T$ the last observed
target value. We reconstruct the median forecast by cumulatively summing the predicted median increments. The remaining quantiles are obtained by aggregating their offsets from the median using a scaled root sum of squares:
\[
  \hat{y}_{q,t}
  = y_T + \sum_{s=1}^{t} m_s
  + \operatorname{sign}\!\left(\sum_{s=1}^{H} o_{q,s}\right)
    \sqrt{\kappa_d}\; t^{\gamma_d-1/2}
    \sqrt{\sum_{s=1}^{t} o_{q,s}^{2}}.
\]
For each quantile, we take the sign of its summed offsets from the median over the full
prediction horizon.
The scale $\kappa_d$ controls the overall width of the quantile bands, while the exponent $\gamma_d$ determines how their width scales with the forecast horizon. Setting $\gamma_d=1/2$ and $\kappa_d=1$ would
leave the root-sum-of-squares envelope unscaled.

\paragraph{Quantile sorting.}
\emph{Also used by: TimesFM-3 and TimeH2O;
available in the current TiRex-2 implementation.}

We sort the predictions at each horizon step into ascending 
order across the model's full quantile grid.
Sorting is performed within each sign-flip branch, before
averaging and selecting the benchmark's quantiles.

\paragraph{Sign-flip augmentation.}
\emph{Also used by: TiRex-2, TimesFM-3, the $t_0$ family, and Timestral. Architecturally enforced by TimeH2O.}

We average the forecast of the raw input with a forecast
obtained by negating its targets and all covariates.
The negated branch uses the same detection, differencing,
integration, and sorting procedures.
Mapping its predictions back to the original scale negates
the predicted values and maps quantile levels
$q\leftrightarrow1-q$.

More precisely, let $F_{q,t}(\mathbf{x})$ denote the forecast
after these operations for input $\mathbf{x}$, where
$\mathbf{x}$ includes the target and encoded covariate values.
The averaged forecast is
\[
  \hat{y}_{q,t}
  = \frac12
    \left[
      F_{q,t}(\mathbf{x})
      - F_{1-q,t}(-\mathbf{x})
    \right].
\]

\paragraph{Nonnegative clipping.}
\emph{Also used by: TimesFM-3, $t_0$-gamma, Timestral, and TimeH2O.}

After sign-flip averaging and selection of the benchmark's
quantile levels, we enforce nonnegativity for targets whose
observed histories contain no negative values.
For each target with exclusively nonnegative observations, we replace
each forecast quantile by
\[
  \hat{y}_{q,t}
  \leftarrow \max\!\left(0,\hat{y}_{q,t}\right).
\]
We check each target from its
original, untransformed history. Targets with negative
observations or no finite observations retain their forecasts.

\paragraph{Integer rounding.}
\emph{Also used by: $t_0$-gamma and TimeH2O.}

After sign-flip averaging and nonnegative clipping,
we round each forecast quantile for targets whose finite
historical observations are all integer-valued:
\[
  \hat{y}_{i,q,t}
  \leftarrow
  \operatorname{round}\!\left(\hat{y}_{i,q,t}\right).
\]

\paragraph{Postprocessing constants.}
\emph{Parameterisation inherited from TiRex-2.}

The core postprocessor has five numerical constants:
the trend window $w$, the trend thresholds $\tau$ and $\rho$,
and the band exponent and scale $\gamma_d,\kappa_d$ of the
differenced-target envelope.
We keep the trend constants of TiRex-2, fit the band constants once
on fev-bench validation forecasts of an 82.5M-parameter development
model, and use the same fixed settings for every ci-proto model.

\subsection{Optimisation and checkpoint selection}

Table~\ref{tab:training-settings} records the optimisation
and execution settings. 

\begin{table}[t]
      \centering
      \small
      \caption{\textbf{Pre-training settings for ci-proto.}
      Size-dependent values follow the order 3.5M, 9.2M, 28.6M, 82.5M, and 167.2M.}
      \label{tab:training-settings}
      \begin{tabular}{p{0.43\linewidth}p{0.49\linewidth}}
          \toprule
          Setting & Value \\
          \midrule
          Optimiser and coefficients &
          AdamW; $\beta_1$ cycled between $0.95$ and $0.85$ by the
          one-cycle schedule ($0.85$ at the peak learning rate),
          $\beta_2=0.999$, $\epsilon=10^{-8}$ \\
          Learning rate and schedule &
          Cosine one-cycle schedule with $5{,}000$ warm-up steps;
          peak LR $1.2\times10^{-3}$ (3.5M--28.6M) and
          $6\times10^{-4}$ (82.5M, 167.2M);
          initial LR $=\mathrm{peak}/50$,
          final LR $=\mathrm{initial}/10^4$ \\
          Effective batch size &
          $256$ sampled time series groups per optimiser step \\
          Training steps per variant &
          $330{,}000; 467{,}500; 700{,}000; 1{,}045{,}000; 1{,}358{,}500$ \\
          Weight decay and gradient clipping &
          Weight decay $0.01$, excluding Mamba parameters
          marked for no decay; elementwise gradient clipping
          to $[-1,1]$ \\
          Numerical precision &
          BF16 mixed precision with FP32 parameters \\
          Training hardware &
          $8$ NVIDIA H100 GPUs with $80$\,GB HBM3 each, for every variant \\
          Checkpoint-selection criterion &
          Final checkpoint at the prescribed training budget \\
          \bottomrule
      \end{tabular}
  \end{table}
  
\section{Evaluation details and interpretation}
\label{app:evaluation}

\subsection{Interpretation of the Pareto frontiers}

For a fixed forecasting metric, let $e(m)$ denote the
aggregate error of model $m$ and let $c(m)$ denote either
its parameter count or its reported median
runtime. A model $m$ is Pareto dominated if another evaluated
model $m'$ satisfies
\begin{equation}
    e(m') \leq e(m),
    \qquad
    c(m') \leq c(m),
\end{equation}
with at least one of these a strict inequality.
In other words, a model is Pareto dominated if another model can strictly improve on one of $e$ or $c$ without sacrificing the other.
The Pareto frontier consists of the non-dominated models.

We construct these frontiers separately for parameter
count and runtime, and separately for SQL and MASE.
Frontier membership therefore describes a particular
two-dimensional comparison among the evaluated models.
It does not imply that a model has the lowest error
regardless of cost, or that its aggregate advantage
holds on every individual forecasting task.

\subsection{Runtime comparisons}

We benchmark all selected models, including the baselines and
ci-proto variants, locally on the same hardware. Evaluation and
timing experiments use a single NVIDIA H100 GPU with $80$\,GB HBM3
memory on a host with two Intel Xeon Platinum 8480+ CPUs.
Inference uses FP32 with no BF16 autocasting.
The runtime plots report the median runtime per $100$ forecasts
measured in this local evaluation setup for every model.

\section{MASE and skill score results}
\label{app:mase_skill}

\begin{figure}
    \centering
    \includegraphics[width=\linewidth]{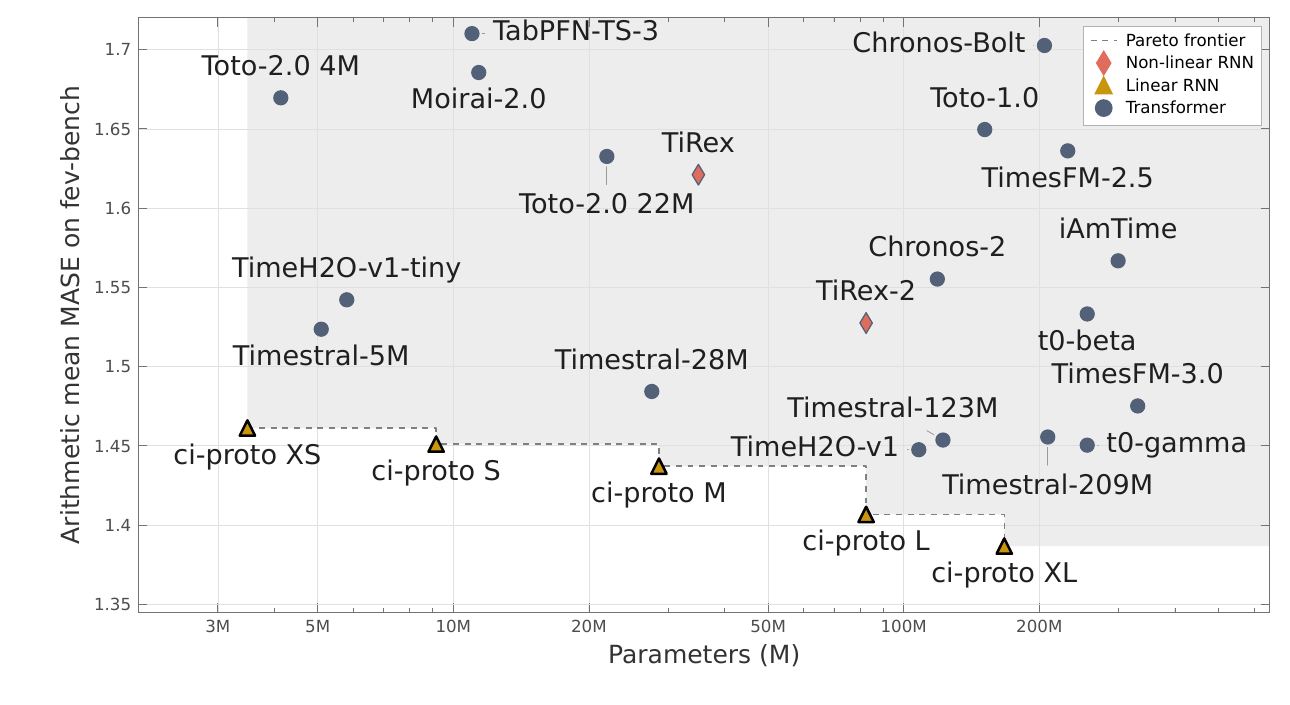}
    \caption{
        \textbf{Mean absolute scaled error (MASE) on fev-bench versus parameter count.} Lower MASE is better. The
        horizontal axis is in logarithmic scale. The dashed line marks the Pareto frontier formed by the five ci-proto
        variants.
    }
    \label{fig:pareto-mase-params}
\end{figure}

\begin{figure}
    \centering
    \includegraphics[width=\linewidth]{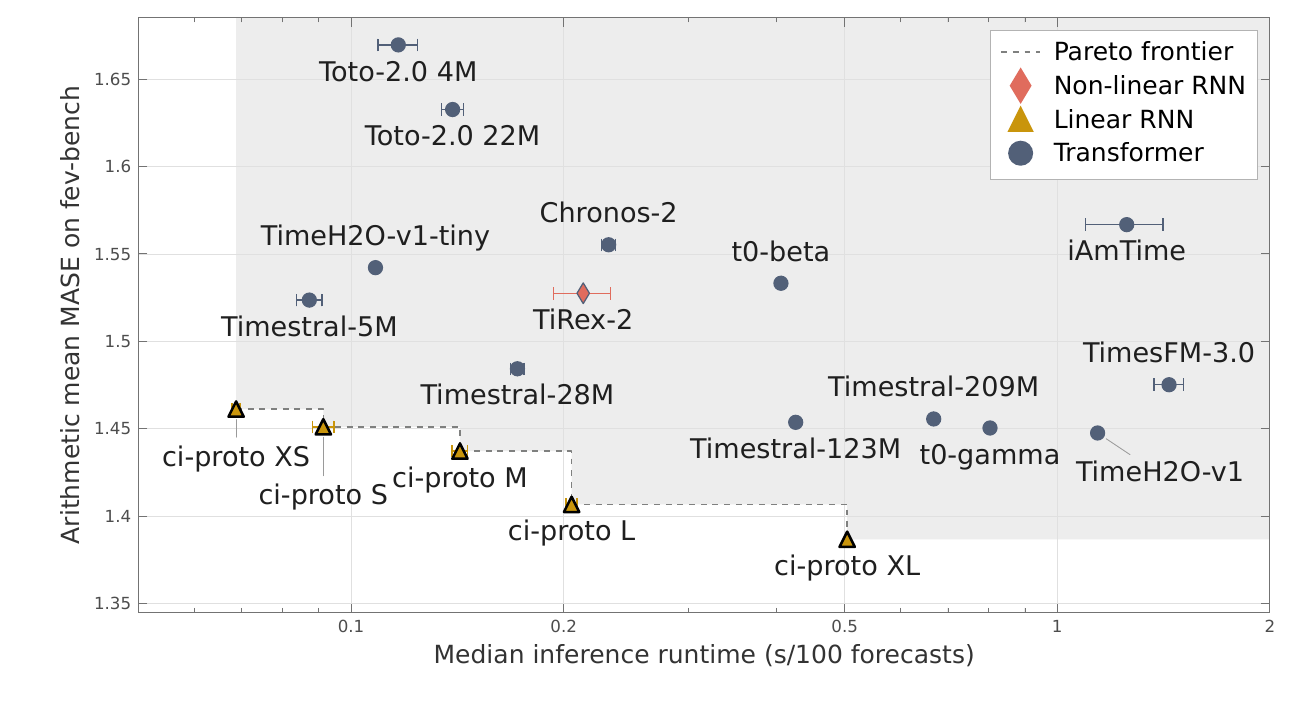}
    \caption{
        \textbf{Mean absolute scaled error (MASE) on fev-bench versus median runtime per 100 forecasts for selected leaderboard models.} We benchmark all models on identical
        hardware and under the same execution settings, since self-reported
        leaderboard timings are not directly comparable. We report the
        mean and standard deviation across five runs. Lower MASE is better. The horizontal axis is in logarithmic scale. The dashed line marks the Pareto frontier formed by the five ci-proto
        variants.
    }
    \label{fig:pareto-mase-runtime}
\end{figure}

Figures~\ref{fig:pareto-mase-params} and~\ref{fig:pareto-mase-runtime} report mean MASE across the $100$ tasks in \texttt{fev-bench}~\citep{shchur2026fevbench}. Unlike SQL, which evaluates multiple predictive quantiles, MASE assesses the predictive median. Mean MASE decreases consistently from ci-proto~XS to ci-proto~XL and all five variants form the Pareto frontiers versus parameter count and measured runtime. This means that every evaluated baseline is dominated in both comparisons, and ci-proto~XL achieves the lowest mean MASE of all models.

\begin{figure}
    \centering
    \includegraphics[width=\linewidth]{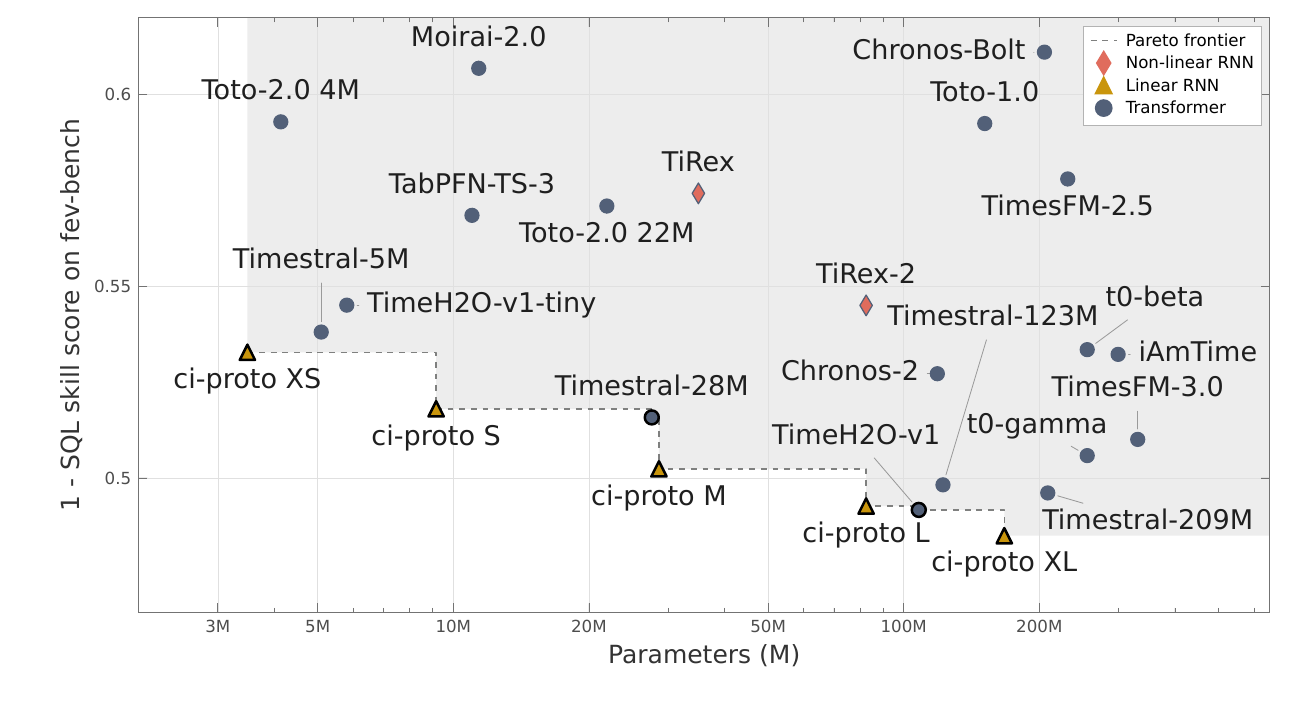}
    \caption{
        \textbf{One minus the skill score for the scaled quantile loss (SQL) on fev-bench versus parameter count.} Lower values are better. The
        horizontal axis is in logarithmic scale. The dashed line marks the Pareto frontier where all five ci-proto variants lie.
    }
    \label{fig:pareto-sql-skill-score-params}
\end{figure}

\begin{figure}
    \centering
    \includegraphics[width=\linewidth]{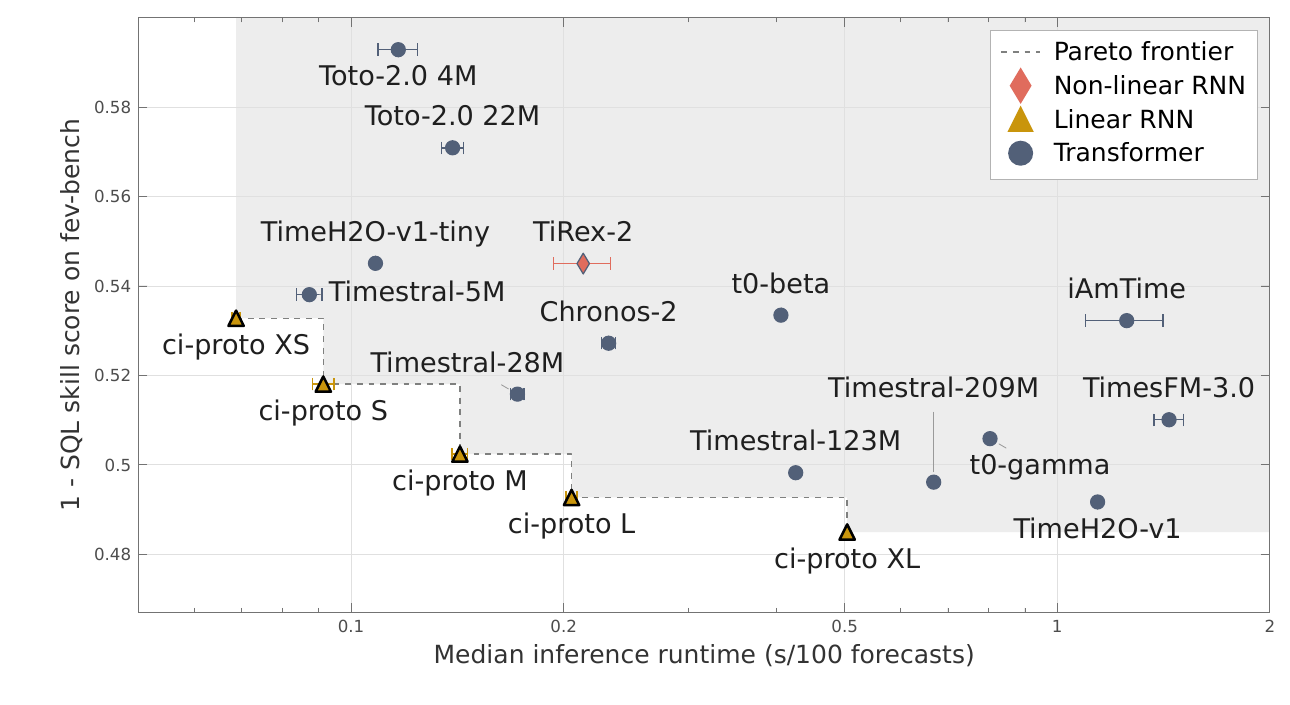}
    \caption{
        \textbf{One minus skill score for the scaled quantile loss (SQL) on fev-bench versus median runtime per 100 forecasts for selected leaderboard models.} We benchmark all models on identical
        hardware and under the same execution settings, since self-reported
        leaderboard timings are not directly comparable. We report the
        mean and standard deviation across five runs. Lower values are better. The horizontal axis is in logarithmic scale. The dashed line marks the Pareto frontier formed by the five ci-proto
        variants.
    }
    \label{fig:pareto-sql-skill-score-runtime}
\end{figure}

\begin{figure}
    \centering
    \includegraphics[width=\linewidth]{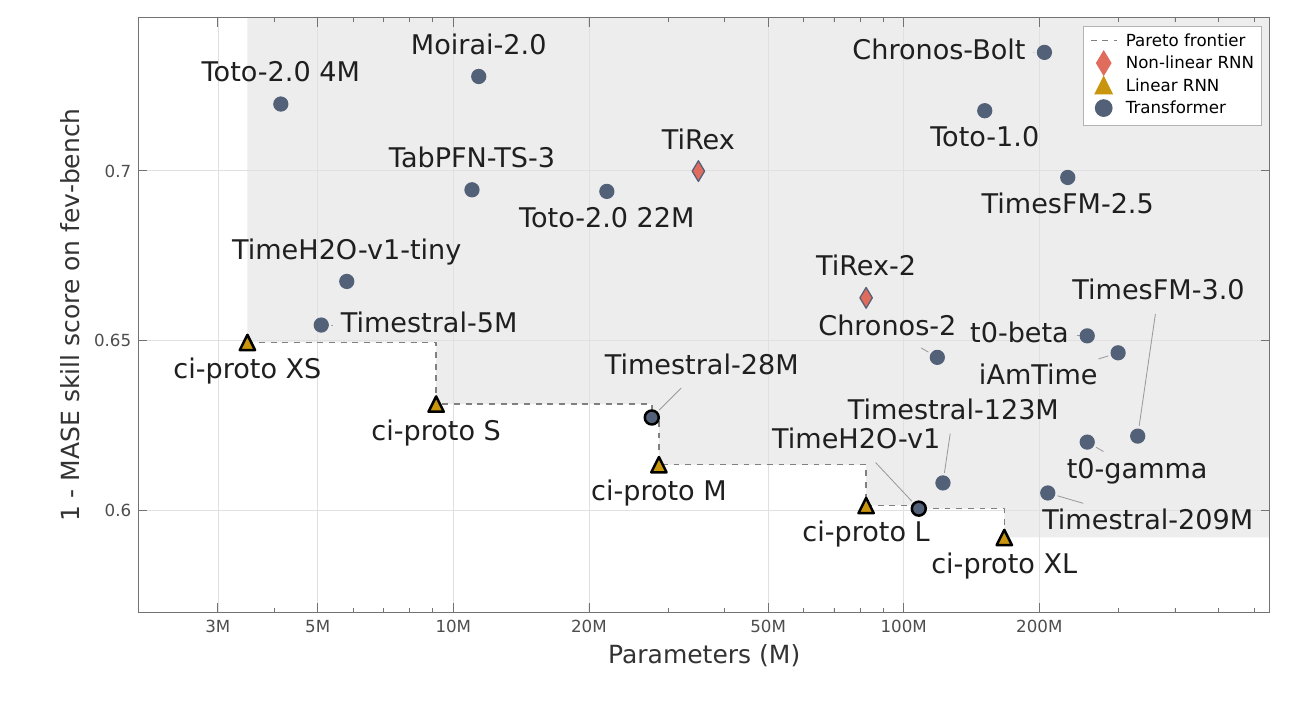}
    \caption{
        \textbf{One minus the skill score for the mean absolute scaled error (MASE) on fev-bench versus parameter count.} Lower values are better. The
        horizontal axis is in logarithmic scale. The dashed line marks the Pareto frontier where all five ci-proto
        variants lie.
    }
    \label{fig:pareto-mase-skill-score-params}
\end{figure}

\begin{figure}
    \centering
    \includegraphics[width=\linewidth]{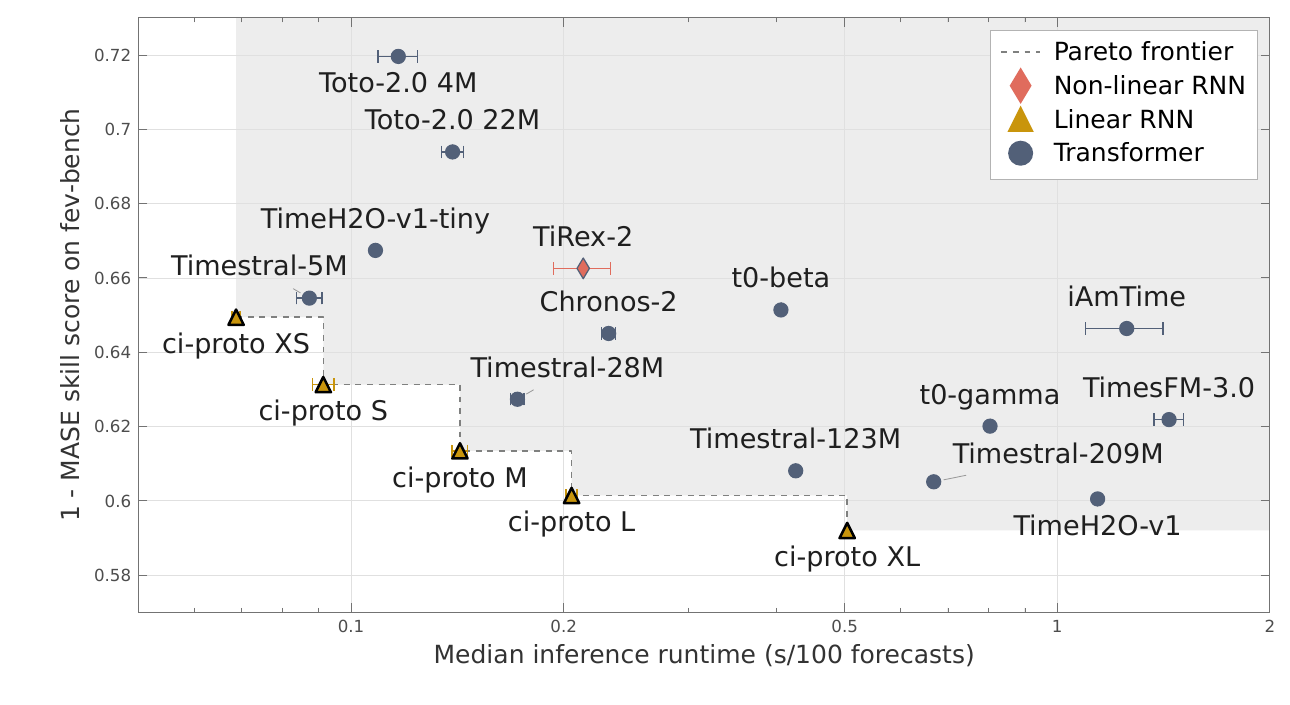}
    \caption{
        \textbf{One minus skill score for mean absolute scaled error (MASE) on fev-bench versus median runtime per 100 forecasts for selected leaderboard models.} We benchmark all models on identical
        hardware and under the same execution settings, since self-reported
        leaderboard timings are not directly comparable. We report the
        mean and standard deviation across five runs. Lower values are better. The horizontal axis is in logarithmic scale. The dashed line marks the Pareto frontier formed by the five ci-proto
        variants.
    }
    \label{fig:pareto-mase-skill-score-runtime}
\end{figure}

Figures~\ref{fig:pareto-sql-skill-score-params}, \ref{fig:pareto-sql-skill-score-runtime}, \ref{fig:pareto-mase-skill-score-params}, and~\ref{fig:pareto-mase-skill-score-runtime} report one minus the SQL and MASE skill scores relative to the seasonal naive baseline, so lower values indicate better performance. All five ci-proto variants remain on the corresponding Pareto frontiers, and ci-proto~XL achieves the best skill scores for both metrics.

% MASE plots

% SQL skill score plots

% MASE skill score plots

\section{Long-range recall}
\label{app:long_range}

For Figure~\ref{fig:long_range_recall}, we fine-tune ci-proto~XL on contexts
of up to $524{,}288$ steps. First, we train for four stages of $4{,}000$, $4{,}000$, $2{,}000$
and $2{,}000$ steps, then for $500$ steps on $1{,}048{,}576$-step
synthetic series alone, and finally for $1{,}000$ steps on its original
training data at a lower learning rate.
In the last three of the four initial stages, about a quarter of the training
series are synthetic series, in which a future-known flag
marks rare events that change the target in a fixed way for each series.
The fine-tune lowers the \texttt{fev-bench} SQL skill score by about 
$1.2$ percentage points.

\begin{figure}
    \centering
    \includegraphics[width=\linewidth]{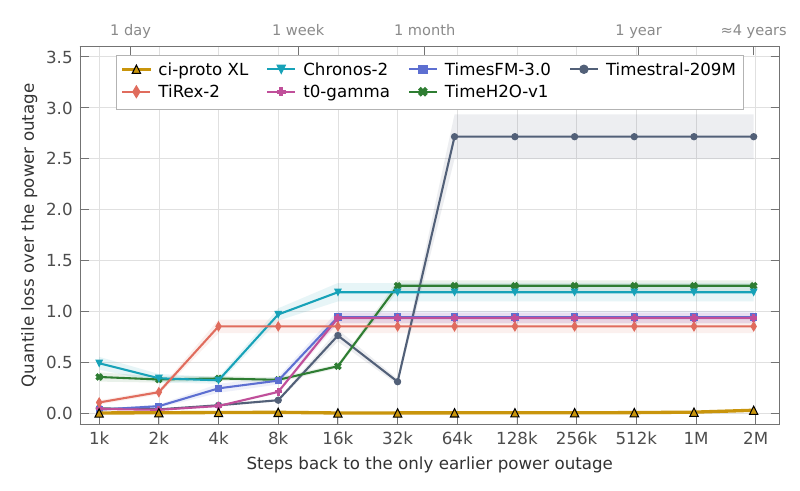}
    \caption{
        \textbf{Quantile loss over a scheduled power outage for models with their released context cap.} The outage has a
        single earlier occurrence at the given number of steps back as shown on the x-axis. 
        Each model receives the smallest of power-of-two context $2^k \geq 2{,}048$ steps that
        contains the earlier outage, or all $2{,}048{,}000$ steps of history if needed.
        ci-proto~XL and the uncapped baselines read the whole context, with all
        other settings matching each model's \texttt{fev-bench} submission.
        The curves show the mean and standard deviation over 40 forecast windows (lower is better). Here each step is one minute, and the top axis gives the
        steps as units of time.
    }
    \label{fig:long_range_recall_capped}
\end{figure}

Figure~\ref{fig:long_range_recall_capped} repeats the scheduled power-outage application with
each baseline as released. This means that each model reads at most a fixed number of recent steps:
$2{,}048$ for TiRex-2, $8{,}192$ for Chronos-2 and t0-gamma,
$15{,}360$ for TimesFM-3.0, $16{,}384$ for TimeH2O-v1 and
$32{,}768$ for Timestral-209M. ci-proto~XL reads the whole context.
As released, each baseline recalls the outage only within its cap.
Beyond its cap, Timestral-209M's upper quantiles widen sharply when the flag switches on.
Figure~\ref{fig:long_range_recall} in the main body removes the caps.

\newpage

\end{document}